\documentclass[runningheads]{llncs}

\usepackage{eccv}

\usepackage{eccvabbrv}

\usepackage{graphicx}
\usepackage{booktabs}
\usepackage{soul}
\usepackage[accsupp]{axessibility}  % Improves PDF readability for those with disabilities.

\usepackage{hyperref}
\usepackage{multirow}
\usepackage{orcidlink}

\begin{document}

% ---------------------------------------------------------------
% TODO REVIEW: Replace with your title
\title{LaxMotion: Rethinking Supervision Granularity for 3D Human Motion Generation} 

% TODO REVIEW: If the paper title is too long for the running head, you can set
% an abbreviated paper title here. If not, comment out.
\titlerunning{Abbreviated paper title}

% TODO FINAL: Replace with your author list. 
% Include the authors' OCRID for the camera-ready version, if at all possible.
% \author{Sheng Liu\inst{1}\orcidlink{0000-1111-2222-3333} \and
% Yuanzhi Liang\inst{2,3}\orcidlink{1111-2222-3333-4444} \and
% Sidan Du\inst{3}\orcidlink{2222--3333-4444-5555}}
\author{Sheng Liu\inst{1} \and
Yuanzhi Liang\inst{2} \and
Sidan Du\inst{3}}

% TODO FINAL: Replace with an abbreviated list of authors.
\authorrunning{F.~Author et al.}
% First names are abbreviated in the running head.
% If there are more than two authors, 'et al.' is used.

% TODO FINAL: Replace with your institution list.
\institute{Nanjing University, Nanjing, China\\
\email{anderliu@smail.nju.edu.cn}\\
\and
TeleAI, Shanghai, China\\
\email{liangyzh18@outlook.com}\\
% \url{http://www.springer.com/gp/computer-science/lncs} \and
\and
Nanjing University, Nanjing, China\\
\email{coff128@nju.edu.cn}}

\maketitle

\begin{abstract}
  Recent 3D human motion generation models demonstrate remarkable reconstruction accuracy yet struggle to generalize beyond training distributions. This limitation arises partly from the use of precise 3D supervision, which encourages models to fit fixed coordinate patterns instead of learning the essential 3D structure and motion–semantic cues required for robust generalization. To overcome this limitation, we propose LaxMotion, a framework that synthesizes realistic 3D motions without direct 3D pose supervision. Instead of regressing toward exact coordinates, LaxMotion learns 3D motion as a consistent explanation of global trajectories and monocular 2D kinematic cues. We introduce a structured motion factorization together with a reformulated training paradigm under relaxed observability. This design is further supported by relaxed regularization objectives that enforce view-consistent alignment, orientation coherence, and structural stability. Under this relaxed supervision paradigm, LaxMotion generates diverse, temporally coherent, and semantically aligned 3D motions, achieving performance comparable to or surpassing fully 3D-supervised methods. These results indicate that shifting supervision from exact coordinate matching to structural consistency promotes stronger reasoning and improved generalization, offering a scalable and data-efficient paradigm for 3D motion generation.
\end{abstract}

\section{Introduction}
\label{sec:intro}

\begin{figure}[!t]
  \centering
  \includegraphics[width=0.8\columnwidth]{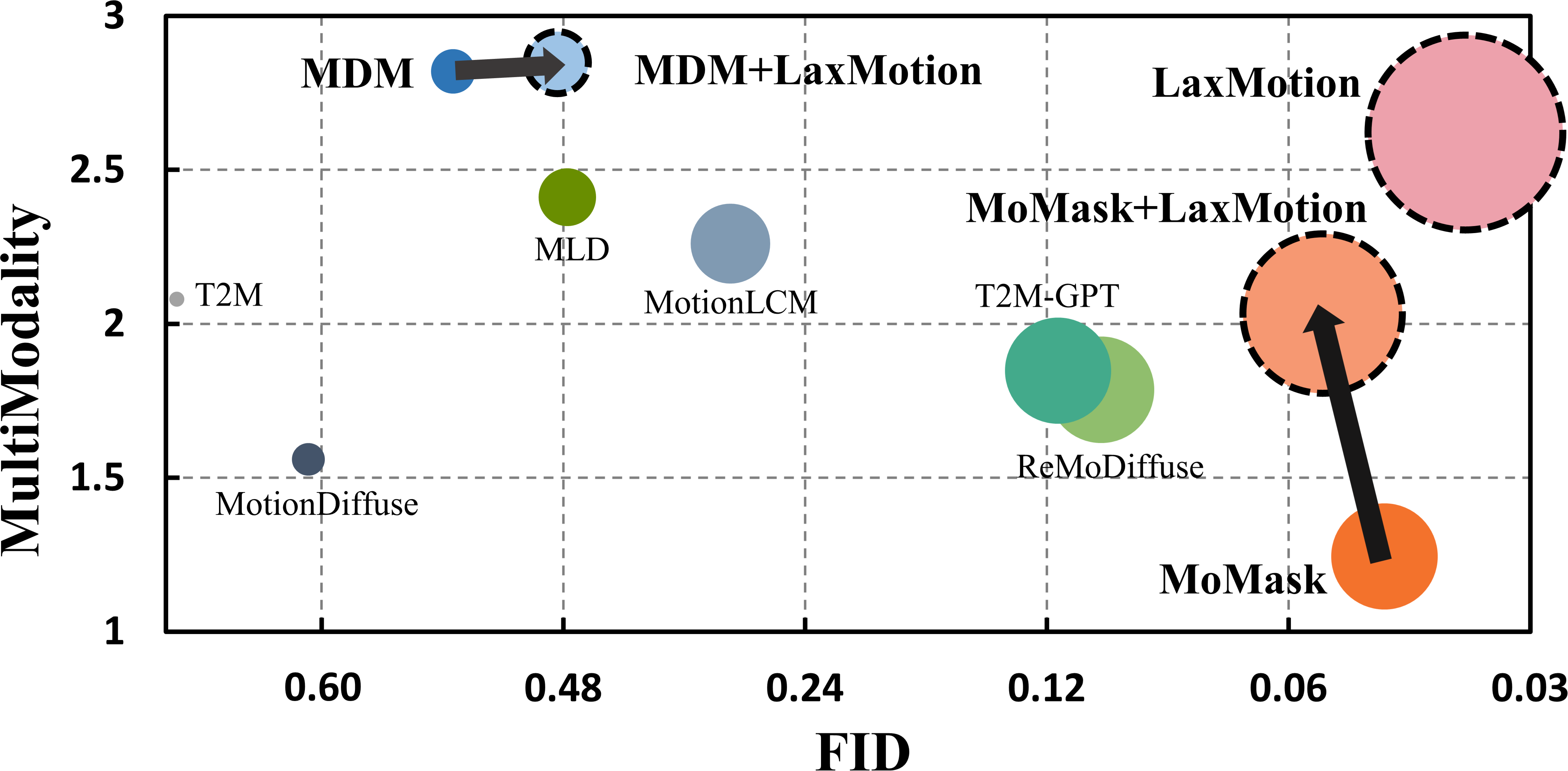}
  \caption{MultiModality-FID comparisons of different methods on HumanML3D dataset. 3D-supervised methods, such as MDM \cite{tevet2023human}, tend to achieve low FID but relatively high MultiModality, whereas MoMask \cite{guo2024momask} exhibits the opposite trend. LaxMotion, trained under a reformulated supervision scheme, relaxes the constraints of precise 3D signals. This encourages the model to learn true motion structure and semantic patterns, enabling it to achieve both high MultiModality and low FID simultaneously. The point size of each method represents its Quality–Multimodality Score (QM), which jointly reflects FID and MultiModality through an integrated measure detailed in the experimental section.}
  % 3D-supervised methods, such as MDM \cite{tevet2023human}, tend to achieve low FID but relatively high MultiModality, whereas MoMask \cite{guo2024momask} exhibits the opposite trend. LaxMotion, trained solely with single-view 2D supervision, aims to relax the constraints imposed by precise 3D signals. This encourages the model to learn true human motion structure and semantic patterns, enabling it to achieve both high MultiModality and low FID simultaneously.}
  \label{fig:intro_cmp}
\end{figure}

% =========

% \section{Introduction}
% \label{sec:intro}

Text-driven 3D human motion generation has progressed rapidly in recent years. Modern diffusion and token-based models can synthesize temporally coherent sequences with strong frame-level fidelity and low reconstruction error on standard benchmarks~\cite{tevet2023human,dai2024motionlcm,zhang2023remodiffuse,guo2024momask,zhang2023generating}. However, high numerical accuracy does not always translate to robust motion understanding. Many systems generalize poorly beyond the training distribution. They often struggle with unseen actions, new subjects, or compositional variations. They can also produce limited diversity for the same input. For instance, masked-token generators such as MoMask~\cite{guo2024momask} may return highly similar samples across repeated generations~\cite{qin2025embracing}. This gap raises a practical question: are we optimizing the right supervision signal for generative motion modeling?

A common design choice in existing methods is to learn a direct mapping from text to 3D joint coordinates~\cite{tevet2023human,dai2024motionlcm,zhang2023remodiffuse,zhang2023generating,guo2024momask}. 3D motion capture provides precise annotations, but it is expensive and often limited in coverage and diversity~\cite{mahmood2019amass,ren2024realistic}. More importantly, coordinate-level supervision is highly over-determined. It encourages exact matching to dataset-specific realizations of motion, including low-level details that are not essential to semantics. As a result, models can achieve strong reconstruction-oriented metrics by fitting the training distribution, while under-learning the invariances that matter for generalization and multi-modality. Recent analyses of VQ-based motion generators support this view. Discrete representations combined with reconstruction-driven objectives tend to reduce diversity and bias generation toward memorized patterns (Fig.~\ref{fig:intro_cmp})~\cite{meng2025rethinking,li2024controlling}.

These considerations suggest that the bottleneck is not only model capacity, but also the granularity of supervision. Text-to-motion generation is inherently one-to-many: a prompt admits multiple valid motions that differ in style, execution, viewpoint, and minor kinematic details. However, coordinate-level 3D supervision converts this one-to-many problem into a point-matching objective. Every training example becomes a single privileged target in $\mathbb{R}^{3J\times T}$, and deviations are penalized even when they remain semantically correct. This mismatch encourages models to fit dataset-specific realizations and low-level coordinate patterns, which can inflate reconstruction-oriented metrics while suppressing diversity and weakening out-of-distribution generalization.

A natural way to better align the learning signal with the generative nature of the task is to supervise motion in a less over-determined space. Single-view 2D kinematics preserves articulation, relative limb dynamics, and temporal ordering, which are central to motion semantics. At the same time, 2D projections discard part of the depth- and camera-dependent details that can dominate coordinate regression. Crucially, a 2D motion sequence does not specify a unique 3D pose sequence; it corresponds to a set of plausible 3D explanations. Training against such a set-valued target relaxes the objective from exact coordinate reproduction to structural consistency. This creates room for multi-modality, while still constraining the motion to remain coherent under time and geometry.

Based on this principle, we train a 3D motion generator without applying losses on ground-truth 3D joint coordinates. We emphasize that this is not a claim that 3D data are unnecessary. On mocap benchmarks, we obtain the required relaxed supervision by projecting available 3D sequences, enabling controlled evaluation. In deployment, the same relaxed signals can be extracted from monocular videos at scale, offering substantially broader coverage. Our supervision combines single-view 2D relative motion with a global trajectory signal, and the model still outputs complete 3D motions. The role of training is to recover 3D as a consistent explanation of these 2D cues under geometric and physical regularization, rather than to memorize 3D coordinates.

% To realize this idea, we introduce \textbf{LaxMotion}, a framework for text-driven 3D motion generation that circumvents dense 3D pose labels by learning from global 3D trajectories and single-view 2D pose cues. LaxMotion has three core components. First, we design a structured motion representation that decomposes motion into a global trajectory and relative limb vectors, defined consistently in both 2D and 3D. This factorization aligns the supervised 2D kinematics with the desired 3D output space. Second, we propose a Motion Lifting Residual Quantized VAE (ML-RQ). It encodes 2D motion into discrete latents and decodes them into 3D motion. This allows the model to learn 3D-consistent codes without 3D pose labels. Third, we introduce a set of 3D-supervision-free regularizations. They include view-consistent projection losses, view-invariant priors induced by random rotations, orientation constraints, and feature-level consistency. Together, these objectives stabilize geometry, improve temporal coherence, and preserve semantic alignment under diverse virtual viewpoints.
To realize this idea, we introduce \textbf{LaxMotion}, a framework that shifts the paradigm of text-to-motion synthesis by rethinking the underlying supervision granularity. Instead of relying on rigid 3D pose-matching, LaxMotion induces high-fidelity 3D structures from global trajectories and monocular 2D kinematic cues through three synergistic strategies. First, we establish a representation reformulation of human motion by decoupling sequences into global trajectories and relative limb vectors, providing a consistent structural basis for coordinate-free pose alignment. Second, we reformulate the training paradigm by relaxing the information source available during learning. Instead of feeding fully specified 3D motion into the generator, we provide only partially observed monocular cues at training time, while still requiring the model to recover complete 3D motion. Finally, we move beyond traditional point-matching by Relaxation Regularizations. They include view-consistent projection losses, cross-view priors induced by random rotations, orientation constraints, and feature-level consistency. Together, these objectives stabilize geometry, improve temporal coherence, and preserve semantic alignment under diverse virtual viewpoints.

Extensive experiments demonstrate that LaxMotion generates natural and diverse 3D motions while remaining faithful to text. Despite using no direct 3D pose-level supervision, it achieves performance competitive with, and in several settings surpassing, strong 3D-supervised baselines on standard benchmarks. These results suggest that relaxing coordinate-level supervision, when paired with appropriate constraints, can improve generalization and diversity in text-to-motion generation.

Our main contributions are:
\begin{itemize}
    \item We identify a limitation of prevailing coordinate-level 3D supervision for generative motion modeling: it can favor dataset-specific fitting and reduce diversity, despite strong reconstruction scores.
    \item We propose \textbf{LaxMotion}, a framework which rethinks supervision granularity by learning from 2D kinematic cues and structural constraints rather than relying on dense 3D pose-level labels.
    % \item We introduce a structured motion factorization, an ML-RQ lifting tokenizer, and 3D-supervision-free regularizations that enforce multi-view geometric stability and temporal coherence.
    \item We introduce a structured motion factorization, a reformulated training paradigm under relaxed observability, and the Relaxation Regularizations that enforce multi-view geometric stability and temporal coherence.
    \item We show that relaxed supervision can yield competitive or superior results to 3D-supervised training, offering a scalable and generalizable alternative supervision strategy.
\end{itemize}

\section{Related Work}
\label{sec:related}
\paragraph{Text-Driven Motion Generation.}
Generating 3D human motion from textual descriptions has become an important research direction, aiming to synthesize semantically consistent and temporally coherent motions from language inputs \cite{chen2023executing, guo2022generating, guo2022tm2t, petrovich2022temos, tevet2023human, guo2024momask, jiang2023motiongpt, zhang2023remodiffuse, liang2024omg, zhang2024motiondiffuse, dai2024motionlcm, wang2025fg}.
Existing approaches can be broadly categorized into diffusion-based and VAE-based methods.
Diffusion models \cite{ho2020denoising, song2020denoising, song2023consistency} generate motions by progressively denoising a latent sequence under a controlled diffusion process. MDM \cite{tevet2023human} first proposed predicting samples instead of noise, enabling geometric constraints during training. ReMoDiffuse \cite{zhang2023remodiffuse} adds a retrieval mechanism to refine denoising, while MotionDiffuse \cite{zhang2024motiondiffuse}, MotionLCM \cite{dai2024motionlcm}, and Fg-t2m++ \cite{wang2025fg} improve semantic alignment and temporal smoothness through enhanced latent diffusion strategies.
VAE-based frameworks such as T2M \cite{guo2022generating} model the probabilistic relationship between text and motion sequences via temporal VAEs \cite{petrovich2021action}. MotionGPT \cite{jiang2023motiongpt} and T2M-GPT \cite{zhang2023generating} combine VAE-based latent motion representations with GPT-like architectures \cite{brown2020language, kojima2022large} to achieve strong semantic consistency. MoMask \cite{guo2024momask} further introduces hierarchical quantization to represent motion as multi-level discrete codes \cite{lee2022autoregressive}.
Despite their success, existing methods predominantly depend on dense 3D motion annotations and rigid coordinate-level supervision. Beyond the high cost of acquiring accurate 3D labels, such tightly constrained supervision can over-determine the learning objective, restricting the model’s flexibility and limiting motion diversity. This motivates the exploration of alternative frameworks that relax supervision while preserving structural coherence, enabling high-quality and semantically faithful motion generation without relying on precise 3D pose annotations.
\paragraph{Weakly-Supervised Motion Generation.}
To alleviate the reliance on dense 3D annotations, weakly-supervised approaches leverage unpaired data or 2D-to-3D lifting priors \cite{chen2020garnet,drover2018can,habekost2020learning,habibie2019wild,kundu2020self,wandt2019repnet,wang2019distill,zanfir2020weakly}. Subsequent works further reduce explicit 3D supervision by enforcing geometric consistency or ambiguity-aware constraints from 2D observations, including adversarial consistency \cite{chen2019unsupervised}, ambiguity mitigation \cite{yu2021towards}, probabilistic flow modeling \cite{wandt2022elepose}, and occlusion-robust extensions \cite{hardy2024links}. More recently, diffusion-based models have been introduced to capture complex motion distributions from in-the-wild videos. For example, MAS \cite{kapon2024mas} adopts multi-view ancestral sampling for cross-view consistency, while MVLift \cite{li2025lifting} progressively reconstructs global 3D trajectories from 2D sequences. Motion-2-to-3 \cite{Guo2025motion2to3} pre-trains a 2D motion generator and subsequently fine-tunes with 3D data to disentangle local and global dynamics. Despite reducing reliance on explicit 3D annotations, existing methods often retain rigid coordinate-level objectives or require partial 3D fine-tuning, limiting motion diversity. Moreover, many approaches remain focused on point-wise alignment or rely on multi-view consistency to enforce constraints, rather than leveraging `relaxed supervision' during training to induce structural consistency. In contrast, LaxMotion revisits supervision granularity and reformulates relaxed constraints, enabling coherent 3D motion synthesis without coordinate-level memorization or 3D pose fine-tuning.

\section{Rethinking Supervision Granularity: LaxMotion}
\label{sec:LaxMotion}
\begin{figure*}[!t]
  \centering
  \includegraphics[width=\textwidth]{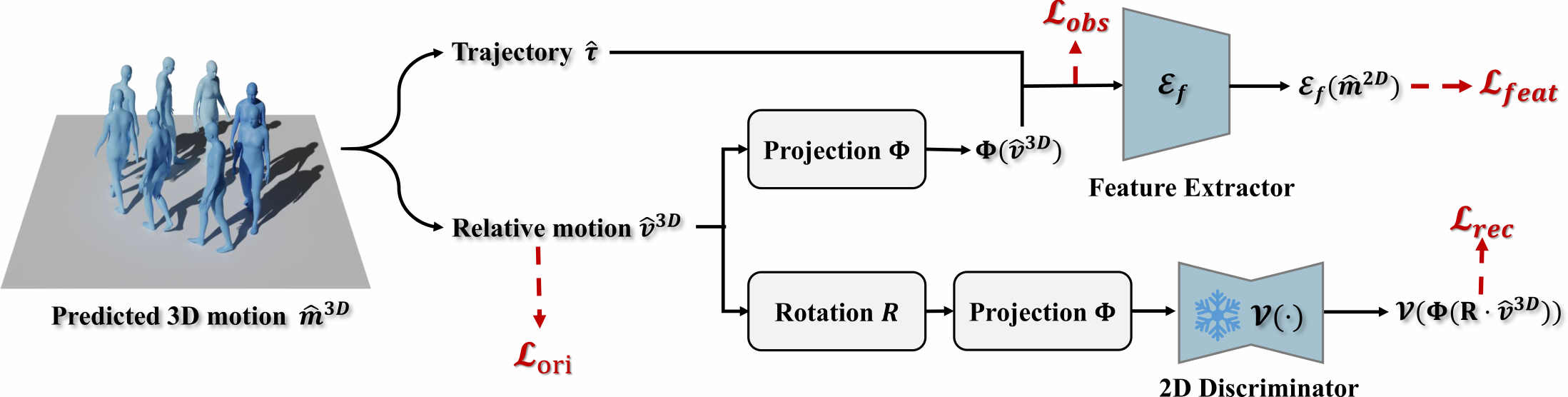}
  \caption{Relaxed supervision for LaxMotion. Instead of rigid coordinate-level point matching, the generator is trained under relaxed supervision using partial 2D motion cues and reprojection consistency. This encourages structurally coherent, view-invariant 3D motion synthesis without explicit 3D pose annotations.}
  \label{fig:method}
\end{figure*}
Our goal is to shift the paradigm of 3D motion generation away from rigid, over-determined coordinate point-matching. To this end, we introduce LaxMotion, a framework that rethinks supervision granularity by learning from 2D kinematic cues and structural constraints rather than relying on dense 3D pose labels. Our approach systematically relaxes the coordinate-level objective through three key components: First, a representation reformulation (Sec.~\ref{subsec: mo_rep}) that transitions the target space from raw 3D points to decoupled kinematic structures; second, a reformulated training paradigm (Sec.~\ref{subsec: generation_architecture}) that relaxes the information source during learning; and finally, a relaxed regularization module (Sec.~\ref{subsec: regularization}) that goes beyond point-matching by enforcing view-invariance and structural coherence, effectively supervising the 3D generation without 3D pose-level annotations.
\subsection{Representation Reformulation: From Points to Structures}
\label{subsec: mo_rep}
% Let $v^{3D}$ represents a set of 3D limb vectors defined as:
% \begin{equation}
%     v^{3D}=\{j^{parent(n)}-j^{child(n)}\}_{n=1}^{K}
% \end{equation}
% where $K$ is the number of limbs. $j^{parent(n)}$ and $j^{child(n)}$ denote the parent and child joints of n-th limb, respectively. We decompose a 3D motion sequence $m_{1:T}^{3D}$ into a global trajectory sequence $\tau_{1:T}$ and a relative motion sequence $v_{1:T}^{3D}$:
% \begin{equation}
%     m_{1:T}^{3D}=\{\tau_{1:T}, v_{1:T}^{3D}\}
% \end{equation}
% This decomposition separates global translation from local articulation, which allows the model to explicitly capture the intrinsic correspondence between 3D relative motion and 2D relative motion. Since our goal is to train the model without any 3D motion supervision, we instead leverage 2D motion data as supervisory signals. Similarly to the representation of 3D relative motion, a 2D relative motion can be represented as a sequence $v^{2D}_{1:T}$, where each $v^{2D}$ is a set of 2D limb vectors describing the relative configuration of body parts. Therefore, the 2D motion can be represented as:
% \begin{equation}
%     m_{1:T}^{2D}=\{\tau_{1:T}, v_{1:T}^{2D}\}
% \end{equation}
Instead of treating human motion as a set of over-determined point-level targets, we reformulate it into a structured representation. We decouple motion into a global trajectory and local limb articulations, which naturally preserves kinematic consistency across spatial dimensions. Specifically, let $v^{3D}$ represents a set of 3D limb vectors defined by the skeletal topology:
\begin{equation}
    v^{3D}=\{j^{parent(n)}-j^{child(n)}\}_{n=1}^{K}
\end{equation}
where $K$ is the number of limbs. $j^{parent(n)}$ and $j^{child(n)}$ denote the parent and child joints of n-th limb, respectively. A complete 3D motion sequence $m_{1:T}^{3D}$ over $T$ frames is then factorized into a global trajectory sequence $\tau_{1:T}$ and a relative motion sequence $v_{1:T}^{3D}$:
\begin{equation}
    m_{1:T}^{3D}=\{\tau_{1:T}, v_{1:T}^{3D}\}
\end{equation}
This decoupling explicitly isolates the root translation from the internal body kinematics. Crucially, this structural factorization bridges the gap between 3D space and 2D projections. By defining motion through relative limb configurations rather than absolute points, we establish a representation that remains mathematically consistent under perspective or orthographic projection. This consistency allows us to fundamentally relax the need for dense 3D coordinate supervision. Instead of penalizing deviations from the exact 3D sequence $m^{3D}_{1:T}$, we drive the learning process using a hybrid kinematic observation, denoted as $m^{obs}_{1:T}$. Symmetrically to the 3D formulation, we substitute the over-determined 3D relative pose with its single-view 2D projection $v^{2D}_{1:T}$, while preserving the global trajectory $\tau_{1:T}$:
\begin{equation}
    m_{1:T}^{obs}=\{\tau_{1:T}, v_{1:T}^{2D}\}
\end{equation}
By structurally aligning the generated $\hat{m}^{3D}_{1:T}$ with its observable cues $m^{obs}_{1:T}$, the model is forced to learn the intrinsic 2D-to-3D geometric correspondence. It recovers 3D realizations as consistent explanations of the observed physical structures, rather than merely memorizing dataset-specific 3D point configurations.

\subsection{Learning from Relaxed Observability}
\label{subsec: generation_architecture}
LaxMotion reformulates the training paradigm by relaxing the information source available during learning. Instead of training the generator with fully specified 3D motion $m^{3D}$ as input, we provide only a partial observation $m^{obs}$ at training time, while still requiring the model to recover complete 3D motion:
\begin{equation}
    \hat{m}^{3D} = \mathcal{G}_\theta(m^{obs})
\end{equation}
Importantly, this observation signal is used only in training and is not required at inference. By reducing the observability of the training input, the learning process no longer depends on explicitly provided dense 3D pose annotations. More fundamentally, by replacing fully determined 3D inputs with incomplete cues, the model is prevented from overfitting to exact coordinate patterns. Instead, it is encouraged to infer coherent 3D structure and motion semantics from limited information. The overall training objective is defined as:
\begin{equation}
    \mathcal{L} = \mathcal{L}_{relax} + \alpha\cdot\mathcal{L}_{prior}
\end{equation}
where $\mathcal{L}_{relax}$ denotes Relaxation Regularization (detailed in Sec.~\ref{subsec: regularization}), which encourages the model to generate 3D motions that are consistent with partial 2D observations. As the core supervisory signal, $\mathcal{L}_{relax}$ guides the model to infer coherent and semantically meaningful 3D motions from limited input information.

The term $\mathcal{L}_{prior}$ serves as a structural regularizer that constrains the solution space of the generator.  In token-based (e.g., VQ-style) architectures, $\mathcal{L}_{prior}$ is instantiated as a commitment loss, which enforces alignment between continuous latent encodings and a predefined discrete codebook distribution, thereby stabilizing the structured latent space. In diffusion-based architectures, however, the latent distribution is already governed by a fixed Gaussian prior, and structural bias is implicitly encoded through the forward diffusion process and its noise scheduling. Consequently, no additional explicit structural regularization is required (i.e., $\alpha=0$), and the optimization objective naturally reduces to minimizing the Relaxation Regularization alone.

\subsection{Beyond Point-Matching: Relaxation Regularization}
\label{subsec: regularization}
In the absence of explicit 3D coordinate ground truth, we relax the supervision granularity through a set of consistency-driven constraints. Instead of performing rigid point-matching, our framework optimizes the model by enforcing structural, view-invariant, and physical coherence.
\paragraph{View-Consistent Structural Regularization.} To anchor the generated 3D motion $\hat{m}^{3D}=\{\hat{\tau}, \hat{v}^{3D}\}$ to the observed reality, we first enforce a structural alignment regularization. Specifically, we utilize a weak-perspective projection operator $\Phi(\cdot)$ to map the generated 3D relative limb vectors $\hat{v}^{3D}$ back to the 2D observation space. The structural loss is formulated as:
\begin{equation}
\mathcal{L}_{obs} = \Vert \Phi(\hat{v}^{3D}) - v^{2D} \Vert_2^2 + \Vert \hat{\tau} - \tau \Vert_2^2
\end{equation}
where $\tau$ and $v^{2D}$ are the observed trajectory and 2D relative pose cues from $m^{obs}$, respectively. This term ensures that the generated 3D structure remains a mathematically valid explanation of the original observed evidence.

\paragraph{Cross-View Plausibility Regularization.} While $\mathcal{L}_{obs}$ ensures 2D alignment from a single viewpoint, it cannot resolve depth ambiguity. We therefore propose a Cross-View Plausibility constraint to induce 3D consistency. We hypothesize that a physically valid 3D motion must yield `natural' 2D projections under any arbitrary rotation $R$. To quantify this `naturalness', we leverage a pretrained and frozen 2D motion discriminator $\mathcal{V}(\cdot)$ (e.g., a 2D relative motion VQ-VAE). The 3D motion is encouraged to maintain high reconstruction fidelity under random projections:
\begin{equation}
\mathcal{L}_{rec} = \Vert \mathcal{V}(\Phi(R \cdot \hat{v}^{3D})) - \Phi(R \cdot \hat{v}^{3D}) \Vert_2^2
\end{equation}
Unlike traditional multi-view supervision methods that necessitate synchronized multi-camera arrays or calibrated pose labels from multiple perspectives, our Cross-View Plausibility constraint operates under a strictly single-view setting. By shifting the requirement from spatial correspondence across physical cameras to distributional consistency across virtual rotations, our approach significantly lowers the barrier for training 3D motion models, enabling high-fidelity synthesis even from monocular `in-the-wild' 2D datasets where multi-view annotations are fundamentally unavailable.

\paragraph{Orientation Regularization.} To enhance physical plausibility, we introduce an Orientation Regularization grounded in the geometric prior that global body orientation and foot direction are intrinsically coupled. Let $\hat{v}_{ori}$ denote the body orientation vector derived from the normalized cross product of hip vectors. Specifically, the forward projection of the foot direction $\hat{v}_{foot}$ onto the body orientation is constrained to be non-negative:
\begin{equation}
    \mathcal{L}_{ori}=\max(0, -\hat{v}_{foot}\cdot\hat{v}_{ori})
\end{equation}
where $\hat{v}_{ori}$ can be computed as the normalized cross product of the left and right hip vectors:
\begin{equation}
    \hat{v}_{ori}=\frac{\hat{v}_{rhip}\times\hat{v}_{lhip}}{\Vert\hat{v}_{rhip}\times\hat{v}_{lhip}\Vert_2}
\end{equation}
\paragraph{Feature Consistency Regularization.} To ensure that the reformulation holds at both the data and feature levels, we introduce Feature Consistency Regularization. The projected motion $\hat{m}^{obs} = \{\hat{\tau}, \Phi(\hat{v}^{3D})\}$ is passed back through one encoder $\mathcal{E}_f(\cdot)$ to ensure alignment between its latent representation and that of the original observation:
\begin{equation}
    \mathcal{L}_{feat}=\Vert\mathcal{E}_f(\hat{m}^{obs})-\mathcal{E}_f(m^{obs})\Vert_2^2
\end{equation}
% Notably, in the token-based pipeline, the encoder $\mathcal{E}_f$ shares parameters with the encoder $\mathcal{E}$ introduced in Sec.~\ref{subsec: generation_architecture}. In the diffusion-based pipeline, the encoder $\mathcal{E}_f$ is learned from the token-based pipeline.
\paragraph{The LaxMotion Regularization.} The final regularization for our supervision is a weighted combination of these structural and prior-based constraints:
\begin{equation}
\mathcal{L}_{relax} = \mathcal{L}_{obs}+\lambda_1\cdot\mathcal{L}_{rec}+\lambda_2\cdot\mathcal{L}_{ori}+\lambda_3\cdot\mathcal{L}_{feat}
\end{equation}
where $\lambda$ are hyperparameters balancing the influence of each geometric constraint.

\section{Experiments}
\label{sec:exp}

\begin{table*}[!t]
\centering
\caption{Quantitative evaluation on the HumanML3D test set. $\pm$ indicates a 95\% confidence interval. `3D Finetune' indicates that LaxMotion is first trained with relaxed supervision to extract features, which are then fused into a 3D-supervised model and fine-tuned. }
\resizebox{\textwidth}{!}{
\begin{tabular}{ccccccccc}
\toprule
\multirow{2}{*}{Supervision} & \multirow{2}{*}{Methods} & \multicolumn{3}{c}{R Precision$\uparrow$} & \multirow{2}{*}{FID$\downarrow$} & \multirow{2}{*}{Diversity$\rightarrow$} & \multirow{2}{*}{MModality$\uparrow$} & \multirow{2}{*}{QM Score$\uparrow$} \\ 
& & Top 1 & Top 2 & Top 3 & & & \\\midrule
& Real & 0.511$^{\pm.003}$ & 0.703$^{\pm.003}$ & 0.797$^{\pm.002}$ & 0.002$^{\pm.000}$ & 9.503$^{\pm.065}$ & - & -\\\midrule
\multirow{9}{*}{3D Motion} & T2M \cite{guo2022generating} & 0.457$^{\pm.002}$ & 0.639$^{\pm.003}$ & 0.740$^{\pm.003}$ & 1.067$^{\pm.002}$ & 9.188$^{\pm.002}$ & 2.090$^{\pm.083}$ & 2.023\\
& MDM \cite{tevet2023human} & 0.320$^{\pm.005}$ & 0.498$^{\pm.004}$ & 0.611$^{\pm.007}$ & 0.544$^{\pm.044}$ & 9.559$^{\pm.086}$ & 2.799$^{\pm.072}$ & 3.795\\
& MLD \cite{chen2023executing} & 0.481$^{\pm.003}$ & 0.673$^{\pm.003}$ & 0.772$^{\pm.002}$ & 0.473$^{\pm.013}$ & 9.724$^{\pm.082}$ & 2.413$^{\pm.079}$ & 3.509\\
& MotionDiffuse \cite{zhang2024motiondiffuse} & 0.491$^{\pm.001}$ & 0.681$^{\pm.001}$ & 0.782$^{\pm.001}$ & 0.630$^{\pm.001}$ & 9.410$^{\pm.049}$ & 1.553$^{\pm.042}$ & 1.957\\
& ReMoDiffuse \cite{zhang2023remodiffuse} & 0.510$^{\pm.005}$ & 0.698$^{\pm.006}$ & 0.795$^{\pm.004}$ & 0.103$^{\pm.004}$ & 9.018$^{\pm.075}$ & 1.795$^{\pm.043}$ & 5.593\\
& T2M-GPT \cite{zhang2023generating} & 0.491$^{\pm.003}$ & 0.680$^{\pm.003}$ & 0.775$^{\pm.002}$ & 0.116$^{\pm.004}$ & 9.761$^{\pm.081}$ & 1.856$^{\pm.011}$ & 5.449\\
& MotionLCM \cite{dai2024motionlcm} & 0.502$^{\pm.003}$ & 0.698$^{\pm.002}$ & 0.798$^{\pm.002}$ & 0.304$^{\pm.012}$ & 9.607$^{\pm.066}$ & 2.259$^{\pm.092}$ & 4.097\\
& MoMask \cite{guo2024momask} & 0.521$^{\pm.002}$ & 0.713$^{\pm.002}$ & 0.807$^{\pm.002}$ & 0.045$^{\pm.002}$ & - & 1.241$^{\pm.040}$ & 5.850\\
& Fg-T2M++ \cite{wang2025fg} & 0.513$^{\pm.002}$ & 0.702$^{\pm.002}$ & 0.801$^{\pm.003}$ & 0.089$^{\pm.004}$ & 9.223$^{\pm.114}$ & 2.625$^{\pm.084}$ & 8.799\\\midrule
\multirow{3}{*}{Relaxed Motion} & LaxMotion (3D Finetune) & \textbf{0.526}$^{\pm.003}$ & \textbf{0.721}$^{\pm.004}$ & \textbf{0.812}$^{\pm.004}$ & \textbf{0.034}$^{\pm.002}$ & \textbf{9.519}$^{\pm.079}$ & 2.529$^{\pm.081}$ & \textbf{13.715}\\
& LaxMotion (MDM-Based) & 0.431$^{\pm.004}$ & 0.624$^{\pm.005}$ & 0.749$^{\pm.006}$ & 0.475$^{\pm.041}$ & 9.523$^{\pm.078}$ & \textbf{2.844}$^{\pm.083}$ & 4.127\\
 & LaxMotion (MoMask-Based) & 0.487$^{\pm.002}$ & 0.681$^{\pm.002}$ & 0.780$^{\pm.002}$ & 0.054$^{\pm.003}$ & 9.486$^{\pm.048}$ & 2.046$^{\pm.068}$ & 8.805\\\bottomrule
\end{tabular}}
% \caption{Quantitative evaluation on the HumanML3D test set. $\pm$ indicates a 95\% confidence interval. `S.V.' denotes using single-view 2D motion for supervision. `3D Finetune' indicates that LaxMotion is first trained with 2D supervision to extract features, which are then fused into a 3D-supervised model and fine-tuned. }
\label{table1}
\end{table*}

\begin{table*}[!t]
\centering
\caption{Quantitative evaluation on the KIT-ML test set. $\pm$ indicates a 95\% confidence interval. `3D Finetune' indicates that LaxMotion is first trained with relaxed supervision to extract features, which are then fused into a 3D-supervised model and fine-tuned.}
\resizebox{\textwidth}{!}{
\begin{tabular}{ccccccccc}
\toprule
\multirow{2}{*}{Supervision} & \multirow{2}{*}{Methods} & \multicolumn{3}{c}{R Precision$\uparrow$} & \multirow{2}{*}{FID$\downarrow$} & \multirow{2}{*}{Diversity$\rightarrow$} & \multirow{2}{*}{MModality$\uparrow$} & \multirow{2}{*}{QM Score$\uparrow$}\\ 
& & Top 1 & Top 2 & Top 3 & & & \\\midrule
& Real & 0.424$^{\pm.005}$ & 0.649$^{\pm.006}$ & 0.779$^{\pm.006}$ & 0.031$^{\pm.004}$ & 11.08$^{\pm.097}$ & - & -\\\midrule
\multirow{8}{*}{3D Motion} & T2M \cite{guo2022generating} & 0.370$^{\pm.005}$ & 0.569$^{\pm.007}$ & 0.693$^{\pm.007}$ & 2.770$^{\pm.109}$ & 10.91$^{\pm.119}$ & 1.482$^{\pm.065}$ & 0.890\\
& MDM \cite{tevet2023human} & 0.164$^{\pm.004}$ & 0.291$^{\pm.004}$ & 0.396$^{\pm.004}$ & 0.497$^{\pm.021}$ & 10.85$^{\pm.109}$ & 1.907$^{\pm.214}$ & 2.705\\
& MLD \cite{chen2023executing} & 0.390$^{\pm.008}$ & 0.609$^{\pm.008}$ & 0.734$^{\pm.007}$ & 0.404$^{\pm.027}$ & 10.80$^{\pm.117}$ & 2.192$^{\pm.071}$ & 3.449\\
& MotionDiffuse \cite{zhang2024motiondiffuse} & 0.417$^{\pm.004}$ & 0.621$^{\pm.004}$ & 0.739$^{\pm.004}$ & 1.954$^{\pm.062}$ & \textbf{11.10}$^{\pm.143}$ & 0.730$^{\pm.013}$ & 0.522\\
& ReMoDiffuse \cite{zhang2023remodiffuse} & 0.427$^{\pm.014}$ & 0.641$^{\pm.004}$ & 0.765$^{\pm.055}$ & 0.155$^{\pm.006}$ & 10.80$^{\pm.105}$ & 1.239$^{\pm.028}$ & 3.147\\
& T2M-GPT \cite{zhang2023generating} & 0.416$^{\pm.006}$ & 0.627$^{\pm.006}$ & 0.745$^{\pm.006}$ & 0.514$^{\pm.029}$ & 10.92$^{\pm.108}$ & 1.570$^{\pm.039}$ & 2.190\\
& MoMask \cite{guo2024momask} & 0.433$^{\pm.007}$ & 0.656$^{\pm.005}$ & 0.781$^{\pm.005}$ & 0.204$^{\pm.011}$ & - & 1.131$^{\pm.043}$ & 2.504\\\midrule
\multirow{3}{*}{Relaxed Motion} & LaxMotion (3D Finetune) & \textbf{0.445}$^{\pm.005}$ & \textbf{0.672}$^{\pm.005}$ & \textbf{0.801}$^{\pm.006}$ & \textbf{0.135}$^{\pm.010}$ & 11.01$^{\pm.097}$ & 2.540$^{\pm.080}$ & \textbf{6.913}\\
 & LaxMotion (MDM-Based) & 0.436$^{\pm.004}$ & 0.644$^{\pm.004}$ & 0.757$^{\pm.005}$ & 0.294$^{\pm.019}$ & 10.49$^{\pm.112}$ & \textbf{2.698}$^{\pm.084}$ & 4.976\\
& LaxMotion (MoMask-Based) & 0.413$^{\pm.008}$ & 0.634$^{\pm.007}$ & 0.763$^{\pm.007}$ & 0.248$^{\pm.016}$ & 11.25$^{\pm.109}$ & 2.481$^{\pm.074}$ & 4.982\\\bottomrule
\end{tabular}}
% \caption{Quantitative evaluation on the KIT-ML test set. $\pm$ indicates a 95\% confidence interval. `S.V.' denotes using single-view 2D motion for supervision. `3D Finetune' indicates that LaxMotion is first trained with 2D supervision to extract features, which are then fused into a 3D-supervised model and fine-tuned.}
\label{table2}
\end{table*}

\begin{table}[t]
\centering
\caption{Comparison with 2D-supervised methods on the HumanML3D test set under global and relative motion settings. `Relative' denotes the setting without trajectory.}
% \small
\resizebox{\columnwidth}{!}{
\begin{tabular}{l l c c c c c c}
\toprule
\textbf{Setting} & \textbf{Method} & R-Precision$\uparrow$ & FID$\downarrow$ & MMDist$\downarrow$ & Diversity$\rightarrow$ & MModality$\uparrow$ & QM Score$\uparrow$ \\
\midrule
\multirow{3}{*}{Global}
& MAS \cite{kapon2024mas}             & 0.418 & 11.893 & 5.606 & 6.413 & -- & -- \\
& Motion-2-to-3 \cite{Guo2025motion2to3}    & 0.697 & 0.321  & 3.579 & 9.286 & 1.892 & 3.339 \\
& LaxMotion (MoMask-Based)    & \textbf{0.780} & \textbf{0.054} & \textbf{3.155} & \textbf{9.486} & \textbf{2.046} & \textbf{8.805} \\
\midrule
\multirow{4}{*}{Relative}
& Real             & 0.713 & --     & 3.791 & 7.731 & -- & -- \\
& Motion-2-to-3 \cite{Guo2025motion2to3}   & 0.602 & 0.779  & 4.534 & 8.450 & 1.148 & 1.301 \\
& MoMask \cite{guo2024momask}          & 0.680 & 0.154  & 4.168 & 8.247 & 0.996 & 2.538 \\
& LaxMotion (MoMask-Based)    & \textbf{0.683} & \textbf{0.137} & \textbf{4.056} & \textbf{8.190} & \textbf{1.311} & \textbf{3.542} \\
\bottomrule
\end{tabular}
}
\label{tab:global_relative_comparison}
\end{table}

\paragraph{Datasets.} We evaluate our approach on two major motion-language benchmarks: HumanML3D \cite{guo2022generating} and KIT-ML \cite{plappert2016kit}. The \textbf{HumanML3D} dataset contains 14,616 motion samples from the AMASS \cite{mahmood2019amass} and HumanAct12 \cite{guo2020action2motion} datasets, each paired with three unique text descriptions, totaling 44,970 descriptions. The \textbf{KIT-ML} \cite{petrovich2021action} dataset integrates data from multiple motion capture sources into a unified format, with 3,911 motion sequences and 6,278 text descriptions, providing a smaller benchmark for evaluation.
% \paragraph{Evaluation Metrics.} We adopt several widely used metrics to comprehensively evaluate the quality, diversity, and text-motion alignment of generated motions. (1) Frechet Inception Distance (FID) measures the distance between the feature distributions of generated and real motions. A lower FID indicates higher motion realism. (2) R-Precision evaluates the semantic alignment between text descriptions and generated motions by measuring the probability that the correct text is ranked within the top-$k$ retrieved texts ($k=1,2,3$) for a given motion. (3) Multi-modal Distance (MMDist) represents the average Euclidean distance between motion and text features, serving as another indicator of cross-modal alignment quality. (4) Diversity quantifies the variability and richness of the generated motions across different text inputs. (5) MultiModality (MModality) measures the variance of motions generated from the same text description, reflecting the model’s ability to produce diverse motions for a single prompt. (6) To jointly evaluate overall generation quality and diversity under identical conditions, we introduce the Quality–MModeality Score (QM), defined as:
\paragraph{Evaluation Metrics.} We adopt several standard metrics to evaluate motion quality, diversity, and text–motion alignment. FID measures the distance between generated and real motion feature distributions, where lower values indicate higher realism. R-Precision assesses semantic alignment by checking whether the correct text ranks within the top-$k$ retrieved ($k=1,2,3$). MMDist is the average Euclidean distance between motion and text features, reflecting cross-modal consistency. Diversity quantifies variability across different text inputs, while MultiModality (MModality) measures variance among motions generated from the same text, capturing conditional diversity. To jointly evaluate overall generation quality and diversity under identical conditions, we introduce the Quality–MModeality Score (QM), defined as:
\begin{equation}
    \text{QM Score}=\frac{\text{MModality}}{\sqrt{\text{FID}}}
\end{equation}
A higher QM indicates a better balance between motion quality and diversity, providing a more holistic evaluation of generative performance.
\subsection{Implementation Details.}
\paragraph{Model Details.} For the token-based pipeline, we adopt MoMask \cite{guo2024momask} as the baseline architecture. For the diffusion-based pipeline, we adopt MDM \cite{tevet2023human} as the baseline framework. In both cases, all architectural designs strictly follow their respective baselines to ensure fair comparison. For the VQ-VAE that learns the 2D motion distribution, we use a single codebook while keeping the other configurations identical.
\paragraph{Training Details.} In our experiments, LaxMotion is implemented under two representative generation paradigms. For both paradigms, we set $\lambda_1=\lambda_2=\lambda_3=1$ during training. For the token-based pipeline, we follow MoMask \cite{guo2024momask} in terms of training protocol with the sole exception of setting $\alpha=2$. For the diffusion-based pipeline, we adopt MDM \cite{tevet2023human} as the baseline and keep all training hyperparameters consistent with its original implementation. All experiments are conducted on a single RTX 3090 GPU.
\subsection{Quantitative Results.}
% We report quantitative results on two benchmark datasets. Unlike prior methods \cite{guo2022generating, tevet2023human, chen2023executing, zhang2024motiondiffuse, zhang2023remodiffuse, zhang2023generating, dai2024motionlcm, guo2024momask, wang2025fg} that require 3D motion data, LaxMotion is trained with single-view 2D supervision, highlighting its ability to perform competitively without 3D annotations.
\paragraph{Comparison on HumanML3D.} As the first 3D motion framework trained primarily under 2D pose cues and 3D trajectories, LaxMotion challenges the convention that coordinate-level 3D supervision is indispensable. As shown in Tab.~\ref{table1}, LaxMotion achieves a competitive FID of 0.054, performing on par with fully 3D-supervised SOTA methods despite avoiding the over-determined point-matching objectives typical of coordinate regression. Notably, LaxMotion attains the highest QM score, reflecting its superior ability to navigate the one-to-many nature of text-to-motion generation by balancing fidelity with multimodality. When fused and fine-tuned with 3D-supervised representations \cite{guo2024momask}, the FID further improves to 0.038, establishing a new SOTA while yielding substantial gains in Diversity. Detailed descriptions of the fusion strategy are provided in the supplementary material. This synergy confirms that 2D-derived representations encode essential structural invariances that effectively complement and regularize traditional 3D modeling. Beyond comparison with fully supervised models, LaxMotion significantly outperforms prior 2D-based approaches across all metrics under the global motion setting (Tab.~\ref{tab:global_relative_comparison}). This demonstrates that by explicitly supervising the 3D trajectory alongside 2D kinematics, our framework generates globally coherent trajectories rather than merely lifting isolated poses. Under the more controlled relative motion setting, LaxMotion remains highly competitive, even surpassing the 3D-supervised MoMask \cite{guo2024momask}. This suggests that shifting from exact coordinate reproduction to structural consistency induces a more generalized and expressive motion representation.
\begin{figure*}[!t]
  \centering
  \includegraphics[width=0.95\textwidth]{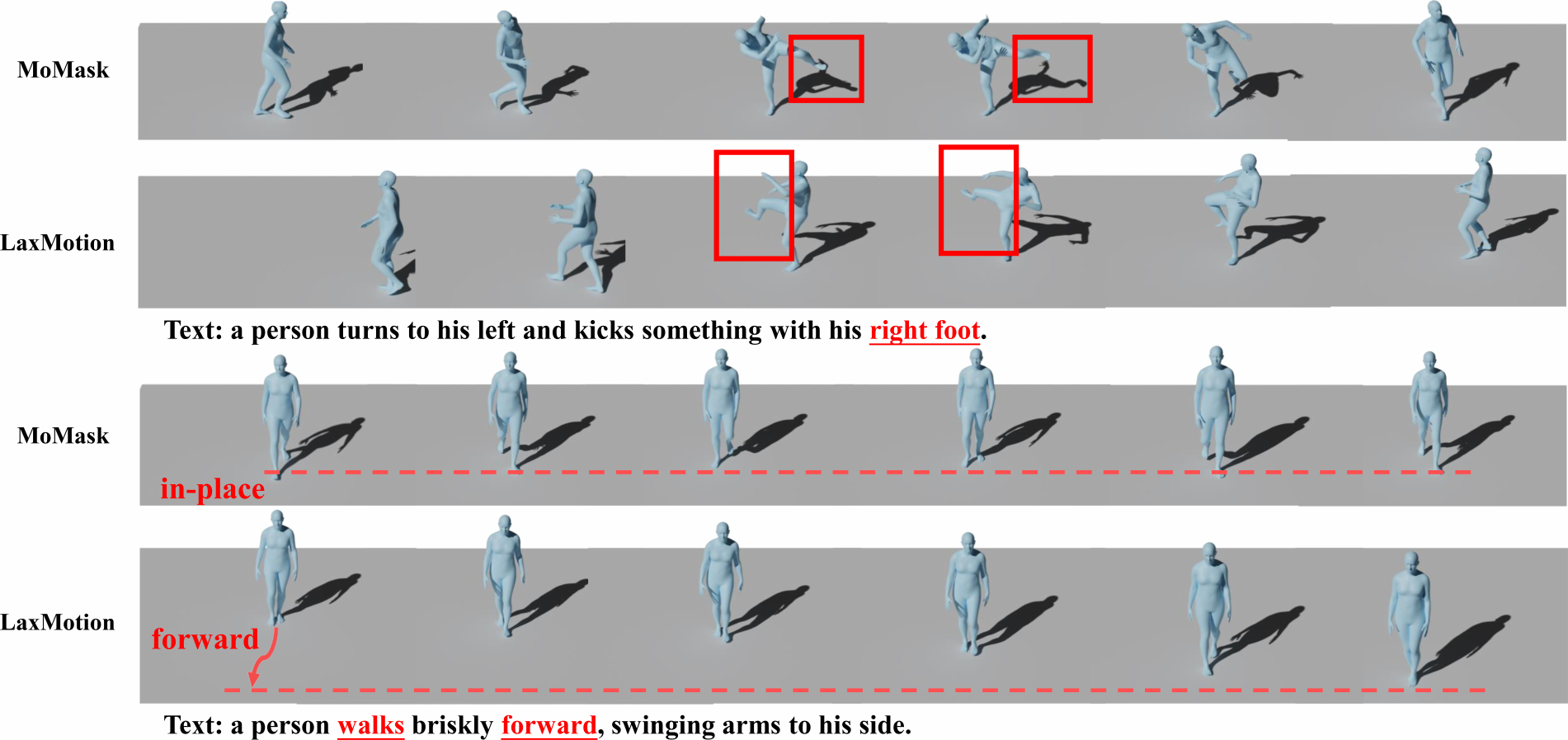}
  \caption{Qualitative comparison on HumanML3D dataset. Compared with MoMask \cite{guo2024momask}, our method demonstrates better generalization and a more accurate understanding of textual semantics, producing motions that more closely align with the given text descriptions.}
  \label{fig:vis1}
\end{figure*}
\begin{figure*}[!t]
  \centering
  \includegraphics[width=0.92\textwidth]{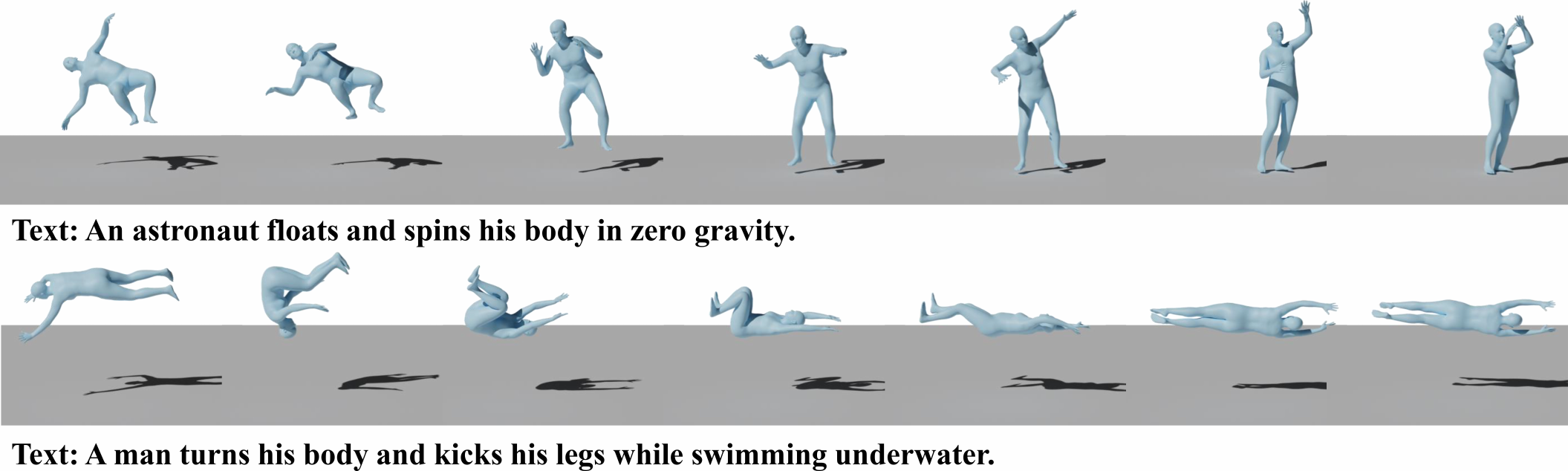}
  \caption{Qualitative Results on In-the-Wild Motions. LaxMotion is capable of learning from and generating motions extracted from in-the-wild 2D videos. We selected several actions that cannot be physically performed or captured with 3D motion sensors under existing conditions, extracted their relaxed motion sequences, and used them for supervision. LaxMotion successfully generates high-quality 3D motions for these challenging scenarios, despite the absence of 3D pose annotations.}
  \label{fig:vis2}
\end{figure*}
\paragraph{Comparison on KIT-ML.} As shown in Tab.~\ref{table2}, we further evaluate LaxMotion on KIT-ML. The results demonstrate that LaxMotion, supervised only by 3D trajectories and 2D pose cues, maintains performance competitive with fully 3D-supervised SOTA methods in FID and R-Precision. Most notably, it achieves the top QM score, reinforcing its superior capacity to balance motion realism with generative diversity—a direct benefit of our relaxed supervision strategy. Furthermore, fusing 2D-derived features with 3D-supervised models yields a new SOTA, confirming that 2D-learned representations capture rich structural and semantic invariants that are often bypassed by traditional 3D coordinate regression. As hypothesized in our Introduction, these findings suggest that while precise 3D pose labels provide strong numerical anchors, they can overconstrain the learning process, leading to a `point-matching' trap. In contrast, the structural abstraction inherent in 2D supervision compels the model to infer underlying kinematic relationships rather than memorizing dataset-specific coordinates. This shift from rote memorization to structural reasoning ultimately results in more robust and diverse motion generation.
\begin{table}[!t]
\centering
\caption{Ablation Study on Relaxation.}
\resizebox{0.85\columnwidth}{!}{
\begin{tabular}{ccccccccc}
\toprule
\multirow{2}{*}{Input} & \multirow{2}{*}{Supervision} & \multicolumn{3}{c}{R Precision$\uparrow$} & \multirow{2}{*}{FID$\downarrow$} & \multirow{2}{*}{Diversity$\rightarrow$} & \multirow{2}{*}{MModality$\uparrow$} & \multirow{2}{*}{QM Score$\uparrow$} \\ 
& & Top 1 & Top 2 & Top 3 & & & & \\\midrule
- & - & 0.511 & 0.703 & 0.797 & 0.002 & 9.503 & - & -  \\\midrule
Full & Relax & 0.511 & 0.706 & 0.806 & 0.040 & 9.684 & 2.240 & 11.200  \\
Relax & Full & 0.507 & 0.688 & 0.791 & 0.048 & 9.805 & 1.288 & 5.879  \\
Relax & Relax & \textbf{0.526} & \textbf{0.721} & \textbf{0.812} & \textbf{0.034} & \textbf{9.519} & \textbf{2.529} & \textbf{13.715} \\\bottomrule
\end{tabular}}
% \caption{Ablation Study on 3D-Free Regularization.}
\label{table_input}
\end{table}

\begin{table}[!t]
\centering
\caption{Ablation Study on Relaxation Regularization.}
\resizebox{0.87\columnwidth}{!}{
\begin{tabular}{ccccccccc}
\toprule
\multirow{2}{*}{Config.} & \multicolumn{3}{c}{R Precision$\uparrow$} & \multirow{2}{*}{FID$\downarrow$} & \multirow{2}{*}{MMDist$\downarrow$} & \multirow{2}{*}{Diversity$\rightarrow$} & \multirow{2}{*}{MModality$\uparrow$} & \multirow{2}{*}{QM Score$\uparrow$} \\ 
& Top 1 & Top 2 & Top 3 & & & & \\\midrule
Real & 0.511 & 0.703 & 0.797 & 0.002 & 2.974 & 9.503 & - & -  \\\midrule
w.o. $\mathcal{L}_{rec}$ & 0.368 & 0.551 & 0.678 & 2.588 & 3.912 & 8.449 & 2.127 & 1.322\\
w.o. $\mathcal{L}_{ori}$ & 0.465 & 0.656 & 0.756 & 0.124 & 3.282 & 9.299 & \textbf{2.225} & 6.319 \\
w.o. $\mathcal{L}_{feat}$ & \textbf{0.501} & \textbf{0.686} & 0.780 & 0.088 & \textbf{3.125} & 9.450 & 2.086 & 7.032  \\
Ours & 0.487 & 0.681 & \textbf{0.780} & \textbf{0.054} & 3.155 & \textbf{9.486} & 2.046 & \textbf{8.805} \\\bottomrule
\end{tabular}}
% \caption{Ablation Study on 3D-Free Regularization.}
\label{table3}
\end{table}
\subsection{Qualitative Results.}
Qualitative comparisons with SOTA 3D-supervised methods are shown in Fig.~\ref{fig:vis1}. Compared to MoMask \cite{guo2024momask}, LaxMotion generates motions more aligned with textual semantics, demonstrating that relaxed supervision improves generalization. Since LaxMotion requires no 3D pose-level annotations, it can be directly applied to in-the-wild video data, learning motions difficult or impossible to capture in 3D. As shown in Fig.~\ref{fig:vis2}, it synthesizes realistic microgravity and underwater motions. Additional results are provided in the supplementary material, highlighting LaxMotion’s potential for scalable and diverse 3D motion generation.
% Qualitative comparisons between LaxMotion and SOTA 3D-supervised methods are presented in Fig.~\ref{fig:vis1}. Compared with MoMask \cite{guo2024momask}, motions generated by LaxMotion exhibit stronger alignment with the intended textual semantics, demonstrating that learning under 2D supervision effectively enhances model generalization. Moreover, since LaxMotion requires no 3D annotations, it can be directly applied to in-the-wild video data, where precise 3D ground truth is unavailable. By training on 2D motion sequences extracted from videos, LaxMotion can learn motions that are extremely difficult to capture in 3D. As shown in Fig.~\ref{fig:vis2}, our framework successfully synthesizes realistic motions in microgravity and underwater swimming where collecting accurate 3D motion data is nearly impossible. More results can be found in supplementary materials. These results highlight LaxMotion’s strong potential for scalable and diverse 3D motion generation beyond the limits of traditional motion capture datasets.
\subsection{Ablation Studies.}
\paragraph{Ablation Study on Relaxation.} Tab.~\ref{table_input} reports the performance of the final model after fine-tuning with features extracted from models trained under varying granularities. The results show that features derived from more relaxed supervision significantly improve performance, indicating that relaxing the supervision granularity enables the model to capture more generalizable motion representations rather than overfitting to exact coordinates. When combined with partial observation during training, this relaxation further enhances the generalization and diversity of motions produced by the feature-fused model, demonstrating the practical benefits of loosening input constraints during training.
\paragraph{Ablation Study on Regularization.} Since LaxMotion relies on 3D trajectories and 2D pose cues, the role of our Relaxation Regularization is pivotal for bridging the 2D-to-3D structural gap. The results in Tab.~\ref{table3} verify the necessity of all three components. Among them, the 2D distribution reconstruction plays the most critical role in improving generation quality. By reconstructing 2D motions of different views, it effectively constrains the multi-view consistency of the generated results and helps the model learn a more accurate 3D spatial structure from purely 2D cues. In addition, both feature consistency regularization and orientation regularization further enhance motion generation quality. The orientation regularization encourages the model to learn consistent and physically plausible body orientations across frames, thereby improving the temporal coherence and realism of generated motions. Meanwhile, feature consistency regularization acts as a latent-space constraint that regularizes the distribution of encoded features, stabilizing the representation learning process and helping the model better disentangle motion semantics from spatial geometry. Collectively, these components enable LaxMotion to learn 3D-aware representations from 2D pose cues.

\begin{table}[!t]
\centering
\caption{Ablation Study on 2D Distribution Learning.}
\resizebox{0.9\columnwidth}{!}{
\begin{tabular}{ccccccccc}
\toprule
\multirow{2}{*}{Config.} & \multicolumn{3}{c}{R Precision$\uparrow$} & \multirow{2}{*}{FID$\downarrow$} & \multirow{2}{*}{MMDist$\downarrow$} & \multirow{2}{*}{Diversity$\rightarrow$} & \multirow{2}{*}{MModality$\uparrow$} & \multirow{2}{*}{QM Score$\uparrow$} \\ 
& Top 1 & Top 2 & Top 3 & & & & \\\midrule
Real & 0.511 & 0.703 & 0.797 & 0.002 & 2.974 & 9.503 & - & - \\\midrule
with VAE & 0.467 & 0.657 & 0.753 & 0.121 & 3.247 & 9.218 & \textbf{2.168} & 6.233\\
with AE & 0.450 & 0.648 & 0.753 & 0.258 & 3.295 & 9.198 & 2.040 & 4.016 \\
% w.o. sharing & 0.477 & 0.660 & 0.759 & 0.120 & 3.281 & 9.230 & 2.147 & 6.198\\
Ours & \textbf{0.487} & \textbf{0.681} & \textbf{0.780} & \textbf{0.054} & \textbf{3.155} & \textbf{9.486} & 2.046 & \textbf{8.805} \\\bottomrule
\end{tabular}}
% \caption{Ablation Study on 2D Motion Distribution Learning and parameter sharing in FCR.}
\label{table4}
\end{table}
\begin{table}[!t]
\centering
\caption{Comparison of token-based reconstructions on various representations.}
\resizebox{0.85\columnwidth}{!}{
\begin{tabular}{cccccccc}
\toprule
\multirow{2}{*}{Config.} & MPJPE$\downarrow$ & \multicolumn{3}{c}{R Precision$\uparrow$} & \multirow{2}{*}{FID$\downarrow$} & \multirow{2}{*}{MMDist$\downarrow$} & \multirow{2}{*}{Diversity$\rightarrow$} \\ 
&(mm) & Top 1 & Top 2 & Top 3 & & \\\midrule
Real & 0 & 0.511 & 0.703 & 0.797 & 0.002 & 2.974 & 9.503 \\\midrule
with joint coordinates & 106.9 & 0.390 & 0.574 & 0.688 & 3.902 & 3.843 & 8.303  \\
Ours & \textbf{56.4} & \textbf{0.487} & \textbf{0.681} & \textbf{0.780} & \textbf{0.054} & \textbf{3.155} & \textbf{9.486} \\\bottomrule
\end{tabular}}
% \caption{Comparison of ML-RQ reconstructions on various representations. `w.j.c' denotes using joint coordinates.}
\label{table5}
\end{table}
\paragraph{Ablation Study on 2D Prior Distribution Learning.} To learn the prior distribution of relative 2D motions, LaxMotion adopts a VQ-VAE architecture for modeling. To verify its necessity, we compare the performance of using VAE and plain AE for distribution learning. As shown in Tab.~\ref{table4}, the results indicate that the performance of VAE is significantly weaker than that of VQ-VAE, while AE performs even worse. This performance gap arises because VQ-VAE, through vector quantization, introduces a discrete representation in the latent space, enabling the model to learn more stable and semantically consistent motion patterns, thus better capturing the structural characteristics of 2D motion distributions. In contrast, the VAE relies on continuous latent variables that often lead to blurred distribution boundaries, while the AE lacks explicit distribution modeling capability and can only perform point-to-point reconstruction, failing to capture the global statistical regularities of motion data. Therefore, VQ-VAE demonstrates clear superiority in learning a high-quality prior of 2D motion distributions, serving as a crucial component that supports ours training paradigm.
% \paragraph{Ablation Study on Parameter Sharing in FCR.} In the token-based pipeline, LaxMotion employs a shared structural encoder for both the initial kinematic encoding and the subsequent feature-level alignment. As shown in Tab.~\ref{table4}, this parameter-sharing strategy improves generation quality: when encoder parameters are not shared, the model’s FID score drops noticeably. This indicates that, within the token-based framework, sharing the encoder provides more consistent and reliable features, helping the model better capture the spatial geometry and dynamic variations of motions.
\paragraph{Ablation Study on LaxMotion Representation.} In our experiments, LaxMotion adopts a limb vector–based representation. This choice is motivated by the fact that limb vectors naturally encode the relative structural relationships of the human skeleton, making it easier for the model to capture geometric constraints and temporal coherence of motions. Additionally, this representation reduces the influence of scale differences between joints on feature learning, thereby improving the stability and accuracy of 3D motion generation. In Tab.~\ref{table5}, we evaluate the impact of different kinematic representations by conducting reconstruction experiments on a token-based VQ-VAE pipeline, including an experiment with joint coordinates directly as the representation. However, this approach ignores the internal skeletal topology, making the model more dependent on absolute positions and more sensitive to noise and scale variations. As a result, the reconstruction quality drops significantly, leading to a substantial performance gap compared to LaxMotion using limb vectors.

\section{Conclusion}
\label{sec:conclusion}
We introduced LaxMotion, a framework that rethinks supervision granularity in text-to-3D motion generation. Instead of regressing toward exact 3D coordinates, LaxMotion learns 3D motion as a consistent explanation of global trajectories and monocular 2D kinematic cues. By shifting from point-level pose matching to relaxed supervision with consistency-driven regularization, our approach mitigates the over-determined nature of dense 3D labels and better aligns the objective with the one-to-many nature of generative motion. Despite not applying direct 3D pose losses, LaxMotion produces natural, diverse, and semantically faithful 3D motions, achieving competitive or superior results compared to fully supervised baselines. These results suggest that structural consistency, rather than exact coordinate memorization, is a more scalable and generalizable principle for 3D motion generation.
\bibliographystyle{splncs04}
\bibliography{main}

@String(ICCV  = {Int. Conf. Comput. Vis.})

@String(ECCV  = {Eur. Conf. Comput. Vis.})

@String(BMVC  = {Brit. Mach. Vis. Conf.})

@String(ICCV  = {ICCV})

@String(ECCV  = {ECCV})

@String(BMVC  =	{BMVC})

@article{ho2020denoising,
  title={Denoising diffusion probabilistic models},
  author={Ho, Jonathan and Jain, Ajay and Abbeel, Pieter},
  journal={Advances in neural information processing systems},
  volume={33},
  pages={6840--6851},
  year={2020}
}

@article{song2020denoising,
  title={Denoising diffusion implicit models},
  author={Song, Jiaming and Meng, Chenlin and Ermon, Stefano},
  journal={arXiv preprint arXiv:2010.02502},
  year={2020}
}

@inproceedings{chen2019unsupervised,
  title={Unsupervised 3d pose estimation with geometric self-supervision},
  author={Chen, Ching-Hang and Tyagi, Ambrish and Agrawal, Amit and Drover, Dylan and Mv, Rohith and Stojanov, Stefan and Rehg, James M},
  booktitle={Proceedings of the IEEE/CVF Conference on Computer Vision and Pattern Recognition},
  pages={5714--5724},
  year={2019}
}

@inproceedings{yu2021towards,
  title={Towards alleviating the modeling ambiguity of unsupervised monocular 3d human pose estimation},
  author={Yu, Zhenbo and Ni, Bingbing and Xu, Jingwei and Wang, Junjie and Zhao, Chenglong and Zhang, Wenjun},
  booktitle={Proceedings of the IEEE/CVF International Conference on Computer Vision},
  pages={8651--8660},
  year={2021}
}

@inproceedings{wandt2022elepose,
  title={ElePose: Unsupervised 3D Human Pose Estimation by Predicting Camera Elevation and Learning Normalizing Flows on 2D Poses},
  author={Wandt, Bastian and Little, James J and Rhodin, Helge},
  booktitle={Proceedings of the IEEE/CVF Conference on Computer Vision and Pattern Recognition},
  pages={6635--6645},
  year={2022}
}

@inproceedings{guo2020action2motion,
  title={Action2motion: Conditioned generation of 3d human motions},
  author={Guo, Chuan and Zuo, Xinxin and Wang, Sen and Zou, Shihao and Sun, Qingyao and Deng, Annan and Gong, Minglun and Cheng, Li},
  booktitle={Proceedings of the 28th ACM International Conference on Multimedia},
  pages={2021--2029},
  year={2020}
}

@inproceedings{petrovich2021action,
  title={Action-conditioned 3d human motion synthesis with transformer vae},
  author={Petrovich, Mathis and Black, Michael J and Varol, G{\"u}l},
  booktitle={Proceedings of the IEEE/CVF International Conference on Computer Vision},
  pages={10985--10995},
  year={2021}
}

@inproceedings{chen2023executing,
  title={Executing your commands via motion diffusion in latent space},
  author={Chen, Xin and Jiang, Biao and Liu, Wen and Huang, Zilong and Fu, Bin and Chen, Tao and Yu, Gang},
  booktitle={Proceedings of the IEEE/CVF Conference on Computer Vision and Pattern Recognition},
  pages={18000--18010},
  year={2023}
}

@inproceedings{guo2022generating,
  title={Generating diverse and natural 3d human motions from text},
  author={Guo, Chuan and Zou, Shihao and Zuo, Xinxin and Wang, Sen and Ji, Wei and Li, Xingyu and Cheng, Li},
  booktitle={Proceedings of the IEEE/CVF Conference on Computer Vision and Pattern Recognition},
  pages={5152--5161},
  year={2022}
}

@inproceedings{guo2022tm2t,
  title={Tm2t: Stochastic and tokenized modeling for the reciprocal generation of 3d human motions and texts},
  author={Guo, Chuan and Zuo, Xinxin and Wang, Sen and Cheng, Li},
  booktitle={European Conference on Computer Vision},
  pages={580--597},
  year={2022},
  organization={Springer}
}

@inproceedings{petrovich2022temos,
  title={TEMOS: Generating diverse human motions from textual descriptions},
  author={Petrovich, Mathis and Black, Michael J and Varol, G{\"u}l},
  booktitle={European Conference on Computer Vision},
  pages={480--497},
  year={2022},
  organization={Springer}
}

@inproceedings{
tevet2023human,
title={Human Motion Diffusion Model},
author={Guy Tevet and Sigal Raab and Brian Gordon and Yoni Shafir and Daniel Cohen-or and Amit Haim Bermano},
booktitle={The Eleventh International Conference on Learning Representations },
year={2023},
url={https://openreview.net/forum?id=SJ1kSyO2jwu}
}

@inproceedings{guo2024momask,
  title={Momask: Generative masked modeling of 3d human motions},
  author={Guo, Chuan and Mu, Yuxuan and Javed, Muhammad Gohar and Wang, Sen and Cheng, Li},
  booktitle={Proceedings of the IEEE/CVF Conference on Computer Vision and Pattern Recognition},
  pages={1900--1910},
  year={2024}
}

@article{jiang2023motiongpt,
  title={Motiongpt: Human motion as a foreign language},
  author={Jiang, Biao and Chen, Xin and Liu, Wen and Yu, Jingyi and Yu, Gang and Chen, Tao},
  journal={Advances in Neural Information Processing Systems},
  volume={36},
  pages={20067--20079},
  year={2023}
}

@inproceedings{zhang2023remodiffuse,
  title={Remodiffuse: Retrieval-augmented motion diffusion model},
  author={Zhang, Mingyuan and Guo, Xinying and Pan, Liang and Cai, Zhongang and Hong, Fangzhou and Li, Huirong and Yang, Lei and Liu, Ziwei},
  booktitle={Proceedings of the IEEE/CVF International Conference on Computer Vision},
  pages={364--373},
  year={2023}
}

@inproceedings{kapon2024mas,
  title={MAS: Multi-view Ancestral Sampling for 3D motion generation using 2D diffusion},
  author={Kapon, Roy and Tevet, Guy and Cohen-Or, Daniel and Bermano, Amit H},
  booktitle={Proceedings of the IEEE/CVF Conference on Computer Vision and Pattern Recognition},
  pages={1965--1974},
  year={2024}
}

@inproceedings{liang2024omg,
  title={Omg: Towards open-vocabulary motion generation via mixture of controllers},
  author={Liang, Han and Bao, Jiacheng and Zhang, Ruichi and Ren, Sihan and Xu, Yuecheng and Yang, Sibei and Chen, Xin and Yu, Jingyi and Xu, Lan},
  booktitle={Proceedings of the IEEE/CVF Conference on Computer Vision and Pattern Recognition},
  pages={482--493},
  year={2024}
}

@article{plappert2016kit,
  title={The kit motion-language dataset},
  author={Plappert, Matthias and Mandery, Christian and Asfour, Tamim},
  journal={Big data},
  volume={4},
  number={4},
  pages={236--252},
  year={2016},
  publisher={Mary Ann Liebert, Inc. 140 Huguenot Street, 3rd Floor New Rochelle, NY 10801 USA}
}

@inproceedings{mahmood2019amass,
  title={AMASS: Archive of motion capture as surface shapes},
  author={Mahmood, Naureen and Ghorbani, Nima and Troje, Nikolaus F and Pons-Moll, Gerard and Black, Michael J},
  booktitle={Proceedings of the IEEE/CVF international conference on computer vision},
  pages={5442--5451},
  year={2019}
}

@article{zhang2024motiondiffuse,
  title={Motiondiffuse: Text-driven human motion generation with diffusion model},
  author={Zhang, Mingyuan and Cai, Zhongang and Pan, Liang and Hong, Fangzhou and Guo, Xinying and Yang, Lei and Liu, Ziwei},
  journal={IEEE transactions on pattern analysis and machine intelligence},
  volume={46},
  number={6},
  pages={4115--4128},
  year={2024},
  publisher={IEEE}
}

@inproceedings{zhang2023generating,
  title={Generating human motion from textual descriptions with discrete representations},
  author={Zhang, Jianrong and Zhang, Yangsong and Cun, Xiaodong and Zhang, Yong and Zhao, Hongwei and Lu, Hongtao and Shen, Xi and Shan, Ying},
  booktitle={Proceedings of the IEEE/CVF conference on computer vision and pattern recognition},
  pages={14730--14740},
  year={2023}
}

@inproceedings{dai2024motionlcm,
  title={Motionlcm: Real-time controllable motion generation via latent consistency model},
  author={Dai, Wenxun and Chen, Ling-Hao and Wang, Jingbo and Liu, Jinpeng and Dai, Bo and Tang, Yansong},
  booktitle={European Conference on Computer Vision},
  pages={390--408},
  year={2024},
  organization={Springer}
}

@article{wang2025fg,
  title={Fg-T2M++: LLMs-augmented fine-grained text driven human motion generation},
  author={Wang, Yin and Li, Mu and Liu, Jiapeng and Leng, Zhiying and Li, Frederick WB and Zhang, Ziyao and Liang, Xiaohui},
  journal={International Journal of Computer Vision},
  pages={1--17},
  year={2025},
  publisher={Springer}
}

@inproceedings{chen2020garnet,
  title={Garnet: Graph attention residual networks based on adversarial learning for 3d human pose estimation},
  author={Chen, Zhihua and Liu, Xiaoli and Sheng, Bing and Li, Ping},
  booktitle={Advances in Computer Graphics: 37th Computer Graphics International Conference, CGI 2020, Geneva, Switzerland, October 20--23, 2020, Proceedings},
  pages={276--287},
  year={2020},
  organization={Springer}
}

@inproceedings{habekost2020learning,
  title={Learning 3D Global Human Motion Estimation from Unpaired, Disjoint Datasets.},
  author={Habekost, Julian and Shiratori, Takaaki and Ye, Yuting and Komura, Taku and Shi, M and Aberman, K and Aristidou, A and Lischinski, D and Cohen-Or, D and Chen, B and others},
  booktitle={BMVC},
  year={2020}
}

@inproceedings{kundu2020self,
  title={Self-supervised 3d human pose estimation via part guided novel image synthesis},
  author={Kundu, Jogendra Nath and Seth, Siddharth and Jampani, Varun and Rakesh, Mugalodi and Babu, R Venkatesh and Chakraborty, Anirban},
  booktitle={Proceedings of the IEEE/CVF conference on computer vision and pattern recognition},
  pages={6152--6162},
  year={2020}
}

@inproceedings{wandt2019repnet,
  title={Repnet: Weakly supervised training of an adversarial reprojection network for 3d human pose estimation},
  author={Wandt, Bastian and Rosenhahn, Bodo},
  booktitle={Proceedings of the IEEE/CVF conference on computer vision and pattern recognition},
  pages={7782--7791},
  year={2019}
}

@inproceedings{wang2019distill,
  title={Distill knowledge from nrsfm for weakly supervised 3d pose learning},
  author={Wang, Chaoyang and Kong, Chen and Lucey, Simon},
  booktitle={Proceedings of the IEEE/CVF international conference on computer vision},
  pages={743--752},
  year={2019}
}

@inproceedings{zanfir2020weakly,
  title={Weakly supervised 3d human pose and shape reconstruction with normalizing flows},
  author={Zanfir, Andrei and Bazavan, Eduard Gabriel and Xu, Hongyi and Freeman, William T and Sukthankar, Rahul and Sminchisescu, Cristian},
  booktitle={Computer Vision--ECCV 2020: 16th European Conference, Glasgow, UK, August 23--28, 2020, Proceedings, Part VI 16},
  pages={465--481},
  year={2020},
  organization={Springer}
}

@inproceedings{hardy2024links,
  title={LInKs" Lifting Independent Keypoints"-Partial Pose Lifting for Occlusion Handling with Improved Accuracy in 2D-3D Human Pose Estimation},
  author={Hardy, Peter and Kim, Hansung},
  booktitle={Proceedings of the IEEE/CVF Winter Conference on Applications of Computer Vision},
  pages={3426--3435},
  year={2024}
}

@inproceedings{habibie2019wild,
  title={In the wild human pose estimation using explicit 2d features and intermediate 3d representations},
  author={Habibie, Ikhsanul and Xu, Weipeng and Mehta, Dushyant and Pons-Moll, Gerard and Theobalt, Christian},
  booktitle={Proceedings of the IEEE/CVF conference on computer vision and pattern recognition},
  pages={10905--10914},
  year={2019}
}

@inproceedings{drover2018can,
  title={Can 3d pose be learned from 2d projections alone?},
  author={Drover, Dylan and MV, Rohith and Chen, Ching-Hang and Agrawal, Amit and Tyagi, Ambrish and Phuoc Huynh, Cong},
  booktitle={Proceedings of the European Conference on Computer Vision (ECCV) Workshops},
  pages={0--0},
  year={2018}
}

@inproceedings{song2023consistency,
  title={Consistency models},
  author={Song, Yang and Dhariwal, Prafulla and Chen, Mark and Sutskever, Ilya},
  booktitle={Proceedings of the 40th International Conference on Machine Learning},
  pages={32211--32252},
  year={2023}
}

@inproceedings{lee2022autoregressive,
  title={Autoregressive image generation using residual quantization},
  author={Lee, Doyup and Kim, Chiheon and Kim, Saehoon and Cho, Minsu and Han, Wook-Shin},
  booktitle={Proceedings of the IEEE/CVF conference on computer vision and pattern recognition},
  pages={11523--11532},
  year={2022}
}

@article{brown2020language,
  title={Language models are few-shot learners},
  author={Brown, Tom and Mann, Benjamin and Ryder, Nick and Subbiah, Melanie and Kaplan, Jared D and Dhariwal, Prafulla and Neelakantan, Arvind and Shyam, Pranav and Sastry, Girish and Askell, Amanda and others},
  journal={Advances in neural information processing systems},
  volume={33},
  pages={1877--1901},
  year={2020}
}

@article{kojima2022large,
  title={Large language models are zero-shot reasoners},
  author={Kojima, Takeshi and Gu, Shixiang Shane and Reid, Machel and Matsuo, Yutaka and Iwasawa, Yusuke},
  journal={Advances in neural information processing systems},
  volume={35},
  pages={22199--22213},
  year={2022}
}

@article{qin2025embracing,
  title={Embracing Aleatoric Uncertainty: Generating Diverse 3D Human Motion},
  author={Qin, Zheng and Wang, Yabing and Yang, Minghui and Zhou, Sanping and Yang, Ming and Wang, Le},
  journal={arXiv preprint arXiv:2508.20604},
  year={2025}
}

@inproceedings{meng2025rethinking,
  title={Rethinking Diffusion for Text-Driven Human Motion Generation: Redundant Representations, Evaluation, and Masked Autoregression},
  author={Meng, Zichong and Xie, Yiming and Peng, Xiaogang and Han, Zeyu and Jiang, Huaizu},
  booktitle={Proceedings of the Computer Vision and Pattern Recognition Conference},
  pages={27859--27871},
  year={2025}
}

@inproceedings{li2024controlling,
  title={Controlling character motions without observable driving source},
  author={Li, Weiyuan and Dai, Bin and Zhou, Ziyi and Yao, Qi and Wang, Baoyuan},
  booktitle={Proceedings of the IEEE/CVF Winter Conference on Applications of Computer Vision},
  pages={6194--6203},
  year={2024}
}

@inproceedings{ren2024realistic,
  title={Realistic human motion generation with cross-diffusion models},
  author={Ren, Zeping and Huang, Shaoli and Li, Xiu},
  booktitle={European Conference on Computer Vision},
  pages={345--362},
  year={2024},
  organization={Springer}
}

@InProceedings{Guo2025motion2to3,
    author    = {Guo, Ruoxi and Pi, Huaijin and Shen, Zehong and Shuai, Qing and Hu, Zechen and Wang, Zhumei and Dong, Yajiao and Hu, Ruizhen and Komura, Taku and Peng, Sida and Zhou, Xiaowei},
    title     = {Motion-2-to-3: Leveraging 2D Motion Data for 3D Motion Generations},
    booktitle = {Proceedings of the IEEE/CVF International Conference on Computer Vision (ICCV)},
    month     = {October},
    year      = {2025},
    pages     = {14305-14316}
}

@inproceedings{li2025lifting,
  title={Lifting motion to the 3d world via 2d diffusion},
  author={Li, Jiaman and Liu, C Karen and Wu, Jiajun},
  booktitle={Proceedings of the Computer Vision and Pattern Recognition Conference},
  pages={17518--17528},
  year={2025}
}
\end{document}